# A fully automated framework for lung tumour detection, segmentation and analysis

Devesh Walawalkar
dwalawal@andrew.cmu.in

## Abstract

Early and correct diagnosis is a very important aspect of cancer treatment. Detection of tumour in Computed Tomography scan is a tedious and tricky task which requires expert knowledge and a lot of human working hours. As small human error is present in any work he does, it is possible that a CT scan could be misdiagnosed causing the patient to become terminal. This paper introduces a novel fully automated framework which helps to detect and segment tumour, if present in a lung CT scan series. It also provides useful analysis of the detected tumour such as its approximate volume, centre location and more. The framework provides a single click solution which analyses all CT images of a single patient series in one go. It helps to reduce the work of manually going through each CT slice and provides quicker and more accurate tumour diagnosis. It makes use of customized image processing and image segmentation methods, to detect and segment the prospective tumour region from the CT scan. It then uses a trained ensemble classifier to correctly classify the segmented region as being tumour or not. Tumour analysis further computed can then be used to determine malignity of the tumour. With an accuracy of 98.14%, the implemented framework can be used in various practical scenarios, capable of eliminating need of any expert pathologist intervention.

## Introduction

According to American cancer society, about 10-20% of all cancer patients (approximately 1.7 millions) are misdiagnosed every year and that at least 40,000 of these patients die just because of it. Moreover, cancer is the second largest cause of death in United states [20]. When a pathologist has on an average 80 Computed Tomography (CT) slides per patient to be analysed, there is a likelihood of misdiagnosing one. Hence there is a need to automate the tumour detection process and in turn eliminate the human error that might creep in.
There is potential for lung cancer to be diagnosed at an earlier stage through the use of screening with low-dose computed tomography, which has been shown to reduce lung cancer mortality by up to 20% [1,15]. Lung CT scan diagnosis is however a highly specialized task requiring expert knowledge. It is quite tricky and time consuming task when done manually. In false negative cases, the tumour manages to miss the human eye as they can be quite similar in texture and size to other particles present in the lungs.
Present work in this topic mainly focuses on qualitative analysis of tumour discovered manually by a pathologist in a single CT scan [2,11,18,27]. The literature lacks a fully automatic framework which provides automatic tumour detection of an entire CT scan series in a single run, prioritizing framework's practicability. Variance in tumour's size, its shape and location limits accuracy of Image segmentation and other methods applied in present work. What distinguishes the proposed segmentation method from present work is the pre-processing done on these images, which greatly helps to improve its accuracy.

Further this paper proposes novel features extracted from the segmented region, which help to distinguish positive tumour cases from the negative ones. This paper aims to provide a practical solution in order to reduce the time and effort required in manual detection process, achieve high accuracy and to completely eliminate human error from the diagnosis.

## Database Collection

The incorporated database consists of CT scan series of 61 patients, each with tumour present in one or more slices. Each series consists of about 70 to 80 CT images. The Database is obtained from the Cancer Imaging Archive (TCIA) public access network [5]. This archive contains collections of CT, PET, MRI scan series of cancer patients, of various body organs like lungs, brain, bladder, prostate etc. This collection is freely available to browse, download, and use for commercial, scientific and educational purposes as outlined in `Creative Commons Attribution 3.0 Unported License'. The database specifically named `Lung CT diagnosis' was created at Moffitt Cancer Centre (Tampa, Florida) [8].
All images are diagnostic contrast enhanced CT scans. The images were retrospectively acquired, to ensure sufficient patient follow-up. Slice thickness is variable between 3 and 6 mm. All images were taken at diagnosis and prior to surgery. Patient data was anonymized and de-identified prior to the analysis. The database includes tumour detection results prepared by experienced radiologists. It provides individual slice numbers of each patient series, where tumour was detected. Part of these results are used to train the classifier incorporated in the framework and remaining are used to test its accuracy.

## Computing Environment

The entire framework is built using MATLAB software package [14]. The testing of this framework and processing of stated results were done on a computing system having an Intel i5 fifth generation processor without any special purpose GPU. With this configuration, the framework took on an average about 1-1.5 minutes to analyse a single patient CT series containing about 70 to 80 slices.

## CT Scan Image Processing

The database contains respective CT images in DICOM (Digital Imaging and Communication in Medicine) format. They are converted into 8 bit gray scale images (Step 1 of Figure 2). The upper and lower 20% of the image is blackened out as it does not contain any useful information or a tumour. It could however mislead the implemented segmentation algorithm as certain isolated elements are present in it. The image is then modified such that pixels having intensity values in range of 110-130, which is the observed tumour intensity range are stepped up to about 210-230 intensity range and all other intensity ranges are stepped down to 10-30 intensity range. This helps to highlight the segmentation target i.e. tumour region in image. It helps the detection algorithm to provide correct and fine segmentation of tumour region. In some lung CT slices, the central part (mainly trachea) of lungs appears isolated from its neighbouring lung wall. This introduces a possibility of the part being falsely segmented by the algorithm. A vertical strip consisting of intensity values in range of 10-30, is superimposed along central region of image. This helps to join central trachea part with its neighbouring wall and correspondingly prevents any false segmentation. An example of resulting processed image can be seen in Step 2 of Figure 2. Closing operation [10,23] is performed to fill in any gaps present inside the tumour. This is followed by Opening operation [10,23] so as to separate any possible tumour clinging loosely to lung wall. The corresponding effect on image can be seen in Step 3 of Figure 2.

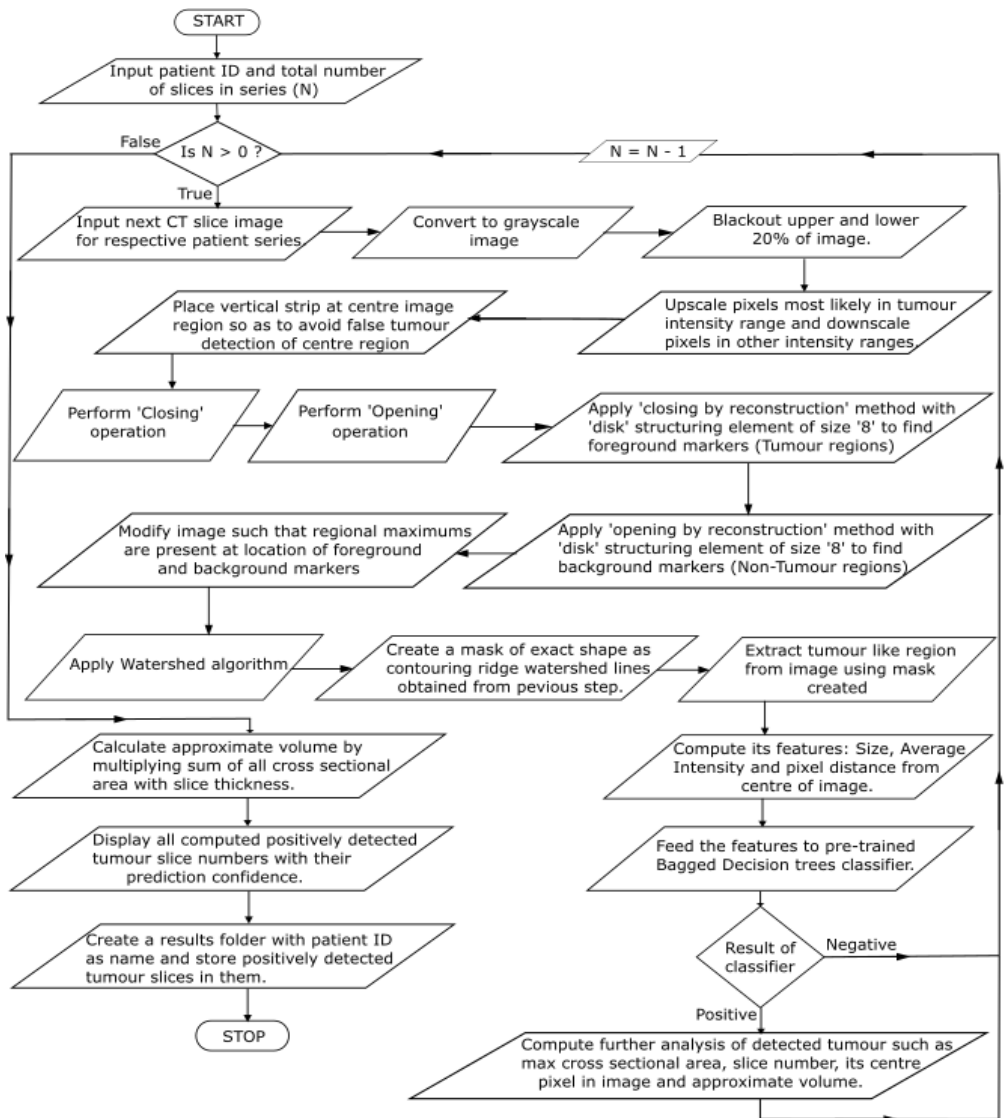

Fig 1. Entire Framework Flowchart

## Tumour segmentation

For segmentation of tumour, marker-controlled watershed algorithm [4,9,16,19,21,24,26,28] is implemented. This method is selected on basis of the fact that lung CT scan consists of a continuous connected mass with only some small parts of it other than the tumour, being separated from the main body. Hence Image segmentation is an ideal method to detect these isolated tumours. Its implementation in medical image analysis has been limited due to possibility of over-segmentation and being noise sensitive. Processing the image to give it apriori knowledge of which region we want to segment, significantly improves the algorithm efficiency.

Since watershed algorithm performs segmentation by drawing ridge lines around local minimums in image, we modify the image such that tumour and non-tumour regions become local minimums. It is done as follows: Firstly, `Opening by reconstruction' technique and subsequently `Closing by reconstruction' technique [7,22] is applied with `disk' structuring element of size 8. Here, a `disk' shape is used as a mask as it somewhat relates to tumour morphology. The size 8 was fixed after experimentation with different sizes. This helps to flatten out the regional maximums present inside image objects pertaining to tumour and non-tumour parts. Further foreground object marker (i.e. Tumour region) is found out by finding regional maximums in image. This process provides both foreground and background markers (i.e. non-tumour region) present in image (Step 4 of Figure 2). They are distinguished using adaptive thresholding technique [13]. The image is modified such that uniform regional minimums are present at the location of computed foreground and background markers in the image. This in effect makes the tumour boundary, a continuous regional maximum. This helps the watershed algorithm to draw ridge lines along tumour's boundary and hence segment it. Regional maximums are computed using `imregionalmax' function present in Image Processing toolbox [14] available in MATLAB software.

Further a mask is created exactly identical to the shape created by the watershed ridge lines (Steps 5 and 7 of Figure 2). The segmented region is then extracted from the image with help of this mask, by keeping all the pixels inside and on the mask same and blacking out the pixels outside the mask (Step 8 of Figure 2).

## Positive classification of Tumour

Classification of segmented region is necessary as certain small non-tumour particles (tertiary bronchi and bronchioles) that are part of the lungs, appear isolated inside certain CT slices. There is a possibility that the implemented segmentation method could falsely segment this as a tumour. To avoid this, novel features are computed from the extracted region. These include the size (number of non-zero pixels) of the region, average intensity of region and pixel distance of region from a vertical line passing through centre of image. The first feature is computed simply by counting the number of non-zero intensity pixels present in segmented region.

Second feature is calculated by dividing sum of intensities of all non-zero pixels in the region by the first computed feature i.e. size. This feature represents the distinguishing property of tumour regions having a specific range of average intensity values only (between 100-120 intensity value), compared to non-tumour ones. Third feature is computed by first finding centre pixel of segmented region. Centre pixel is found out by finding horizontal (X1 and X2) and vertical (Y1 and Y2) extremities of region and then using equation (1).

$$Centre pixel = [(X1 + X2)/2, (Y1 + Y2)/2] \qquad (1)$$

Third feature is computed as the horizontal distance of this centre pixel from vertical centre line in image. Third feature represents the distinguishing property of tumour region being fairly away from centre of image compared to non-tumour ones. These features combined together accurately help to distinguish tumour from other non-tumour lung particles.

A Bagged decision trees classifier [3,6,12,17,25,29] is used here to correctly classify the segmented region. It is implemented with help of Classification learner application available in MATLAB software. Classifier is trained using 15-fold cross validation technique on 60% of patient database. For training purpose, features were computed and fed to the classifier with label `1' for those containing tumour and label `0' for those not containing one. Labelling was done in accordance to individual patient results provided with the database. Bagged Decision trees is chosen over other classification methods owing to its higher classification accuracy (Table 1 and 2), greater control over its model fitting parameters and its ensemble learning feature of using multiple decision tree models having different subsets of the database. This helps the model to better generalize and reduce error further, compared to a single classifier model [6,29].

Table 1 Performance comparison of various classifiers on cross validation set

| Classifier | Percent Accuracy for N-fold cross validation set | | | |
|---|---|---|---|---|
| | N=5 | N=10 | N=15 | N=20 |
| Linear Support Vector Machine | 90.24 | 91.17 | 92.13 | 90.73 |
| Decision Tree | 89.21 | 89.53 | 89.49 | 89.79 |
| K-Nearest Neighbours (K=10) | 92.39 | 93.02 | 90.64 | 92.11 |
| Bagged Decision Trees | 97.48 | 98.34 | 98.72 | 98.26 |
| Gaussian Support Vector Machine | 87.28 | 88.62 | 89.12 | 88.19 |
| Weighted K-Nearest Neighbours | 90.17 | 90.82 | 91.13 | 92.62 |

Table 2 Performance comparison of various classifiers on test set

| Classifier | Percent accuracy for test set |
|---|---|
| Linear Support Vector Machine | 87.96 |
| Decision Tree | 90.60 |
| K-Nearest Neighbours (K=10) | 94.21 |
| Bagged Decision Trees | 98.14 |
| Gaussian Support Vector Machine | 89.12 |
| Weighted K-Nearest Neighbours | 92.44 |

A performance comparison of various classifiers for both cross validation set and test set is presented in table 1 and 2 respectively.

## Tumour analysis

Analysis of positively classified tumour region is further carried out. It includes the maximum cross sectional area of the tumour and its corresponding slice number, its centre location within the image, approximate volume of the entire tumour and percent confidence of it being a tumour (Figure 1). This analysis is computed by finding out the upper and lower bounds of tumour along both X, Y dimensions of the image and then by using equation (1). Cross sectional area is calculated by counting the number of pixels representing the extracted tumour. Approximate volume is computed by summing the cross sectional area of all positively computed tumours and multiplying it by slice thickness of that respective patient series. Maximum cross sectional area is found by comparing areas of all positively detected slices and finding maximum among it.

Table 3 Test set confusion matrix

| Total slices in test set N = 1720 | Predicted Positive | Predicted Negative | Total |
|---|---|---|---|
| Actual Positive | 72 | 1 | 73 |
| Actual Negative | 31 | 1616 | 1647 |
| Total | 103 | 1617 | 1720 |

## Classifier model and flowchart

A sequential flowchart of the proposed framework can be seen in Figure 1. Users simply need to add the file into the working directory of MATLAB. The model named `Classifier' could then be loaded into their workspace. The trained model can be used for prediction using `predict' function provided by Statistics and Machine learning toolbox [14] present in MATLAB software package. Users can use this trained model or train one on their own using appropriate data of segmented region features.

## Results

Implemented framework gives an accuracy of 98.14% on test set, for both the true positive and true negative cases combined. Considering only true positive cases, accuracy is 98.63% and for true negative cases, accuracy is 97.02% (Refer Table 3). Framework accuracy over entire database is 98.40%. The feature value ranges for a positive tumour region learnt by classifier are as follows: 100-120 intensity/pixel for average intensity, 1000-5000 pixels for tumour size and 80-110 pixel distance between tumour centre and vertical central line in image. The step wise processing of CT scan image through the framework can be seen in Figure 2.

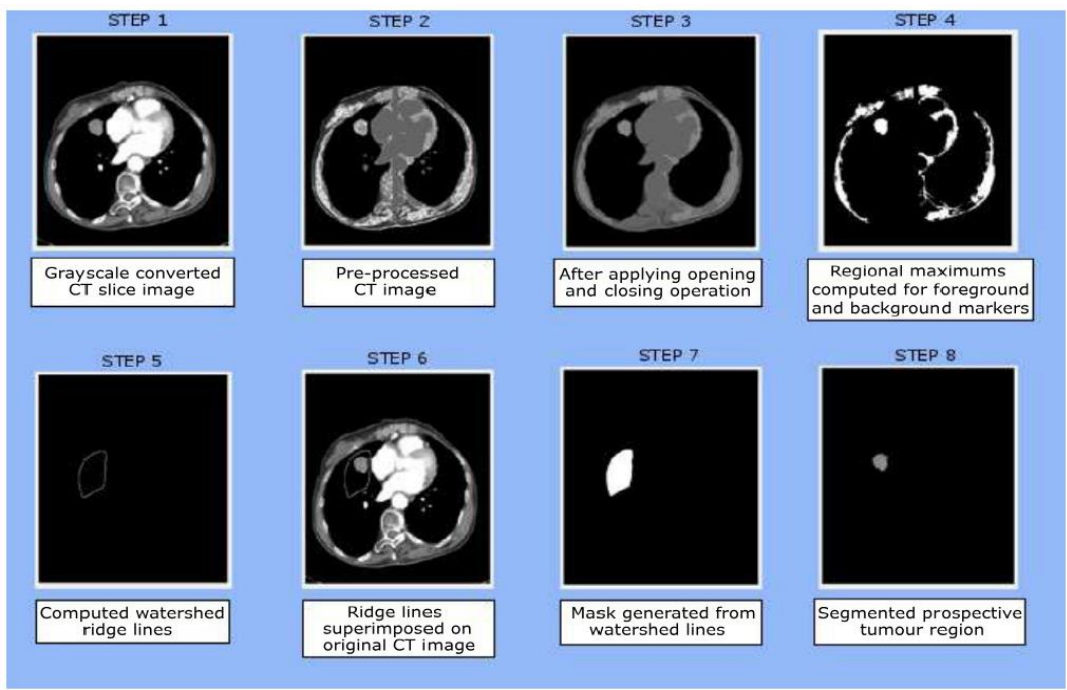

Fig. 2 Step wise processing of CT image while running through framework

## Discussion

As seen from results, an accuracy of 98.14% is good enough to implement this system in practical scenarios. The results also provide evidence for the fact that the three novel tumour features proposed in this paper, together help to classify segmented tumour from segmented lung particles with excellent accuracy. The analysing speed of the framework at about 1-1.5 minutes per series is a considerable improvement over manual speed of an individual pathologist. Computed tumour analysis would help the operating pathologist to gain in depth knowledge of the detected tumour. Furthermore, framework stores the positively detected tumour slices in a run time created folder having the patient ID as its name. An example of images stored in these result folder can be seen in Step 6 of Figure 2. This provides for the pathologist to review and confirm the tumours detected by framework. This serves to add an extra layer of positive tumour confirmation.

Certain limitations of this proposed framework also exist. It includes the fact that only CT images can be processed. There is currently no option for PET and MRI scans. It has been customized to detect tumours in lung region only. Further modifications need to be done for detection in other body

## Conclusion

This automated framework thus reduces human error and provides a fast, efficient and reliable method to diagnose and analyse tumour regions in lung CT scans. Future scope in this work includes working on incorporating the tumour detection in all types of CT scans of other parts of the body. Also a critical aspect to include would be classify the detected tumour with sufficient confidence, whether it is a malignant or a benign tumour. Also the extent (Stage) of a malignant tumour can be incorporated.

## Acknowledgements

Special thanks to Moffitt Cancer Centre (Tampa Florida, US) along with Ms Olya Stringfield, PhD from the `Department of Cancer Imaging and Metabolism' for preparing this database and for providing an indexed table containing positive tumour slice numbers of each patient series. Also special acknowledgment to The Cancer imaging archive (TCIA) for making this database available under public license and for its efficient maintenance.